\documentclass[11pt]{article}

\usepackage[]{acl}

\usepackage{times}
\usepackage{latexsym}
\usepackage{subcaption}
\usepackage{amsmath, amssymb}

\usepackage[T1]{fontenc}

\usepackage[utf8]{inputenc}

\usepackage{microtype}

\usepackage{inconsolata}
\usepackage{tcolorbox}
\usepackage{url}

\usepackage{graphicx}
\usepackage{booktabs}
\usepackage{multirow}
\usepackage{amsmath}
\usepackage{makecell}
\usepackage{rotating}

\usepackage{xcolor}

\title{Steer Model beyond Assistant: Controlling System Prompt Strength via Contrastive Decoding}

\author{Yijiang River Dong$^*$, Tiancheng Hu$^*$, Zheng Hui$^*$, Nigel Collier \\ \{yd358, th656, zh403, nhc30\}@cam.ac.uk \\
         University of Cambridge}

\begin{document}
\maketitle
\def\thefootnote{*}\footnotetext{Equal contribution}\def\thefootnote{\arabic{footnote}}
\begin{abstract}
Large language models excel at complex instructions yet struggle to deviate from their helpful assistant persona, as post-training instills strong priors that resist conflicting instructions. We introduce system prompt strength, a training-free method that treats prompt adherence as a continuous control. By contrasting logits from target and default system prompts, we isolate and amplify the behavioral signal unique to the target persona by a scalar factor $\alpha$. Across five diverse benchmarks spanning constraint satisfaction, behavioral control, pluralistic alignment, capability modulation, and stylistic control, our method yields substantial improvements: up to +8.5 strict accuracy on IFEval, +45pp refusal rate on OffTopicEval, and +13\% steerability on Prompt-Steering. Our approach enables practitioners to modulate system prompt strength, providing dynamic control over model behavior without retraining. Our code will be available at \url{https://github.com/dong-river/persona_cd}.

\end{abstract}

\section{Introduction}

Ask a language model to role-play as a six-year-old child and answer ``What is the melting point of iron'' and it replies, with such precision, ``It's 1,538 degrees Celsius'' (Figure~\ref{fig:main}). 
Instruct it to refuse off-topic queries as a medical assistant, and it cheerfully explains quantum mechanics. 
Modern LLMs can follow remarkably complex instructions, solving multi-step math, summarizing dense legal text, debugging intricate code~\citep{achiam2023gpt, hui-etal-2025-winspot, team2025gemma}, yet they struggle with something far simpler: \emph{not} being helpful.

This paradox reflects a tension inherent to LLM training. Post-training pipelines typically optimize for helpfulness and harmlessness, conditioning the model to be an articulate, knowledgeable, and accommodating assistant~\citep{bai2022constitutional, ouyang2022training}. When applications demand non-standard behavior, e.g. a tutoring bot that intentionally struggles, a specialist agent that must refuse off-topic queries, a pluralistic system representing diverse perspectives, this assistant persona proves difficult to suppress~\citep{miehling_evaluating_2025, kumar2025can, jiang2025artificial}.

Today, to steer a model via natural language, practitioners have effectively a single lever: the system prompt. They can \emph{describe} the persona they want, but they cannot \emph{modulate} how strongly the model should commit to it. While one could fine-tune a separate model for each use case, this is computationally expensive, data-dependent, and model-specific. Activation steering offers an inference-time alternative, yet it remains an experimental technique with a high barrier to entry: it lacks robust standardized tooling, requires white-box access, and faces generalization issues~\cite{tan2024analysing,pres2024towards}. We take a complementary approach: treating the strength of the system prompt itself as an explicit decoding-time control.

We introduce \textbf{system prompt strength} ($\alpha$), a continuous hyperparameter that governs the intensity of system prompt's influence. Our method isolates the specific behavioral "delta" by comparing the logits generated with the system prompt against those from the default system prompt. $\alpha$ acts as the amplification coefficient for this delta: it determines how aggressively we shift the output distribution away from the default prior and toward the target persona. At $\alpha{=}0$, we recover standard decoding; at higher $\alpha$, the assistant prior fades and the persona sharpens. The system prompt becomes not just a description but a \emph{tunable dial}.

We systematically evaluate our approach across five model families and five diverse benchmarks, ranging from instruction following (IFEval~\citep{zhou2023instruction}, OffTopicEval~\citep{lei2025offtopiceval}) to role-playing  and pluralistic alignment (Prompt-Steering~\citep{miehling_evaluating_2025}, Counterfactual Simulation~\citep{kumar2025can}, and {Readability Control).
Our method yields substantial, training-free gains: \textbf{+8.5} strict accuracy on IFEval, \textbf{+45pp} refusal rate on OffTopicEval, and \textbf{+13\%} steerability on Prompt-Steering. These improvements stack with existing prompts and require no additional training. In short: where prompt engineering offers a \emph{description}, we offer a \emph{dial}.

\section{Related Work}

\begin{figure*}[!ht]
    \centering
    \vspace{-6mm}
    \includegraphics[width=0.92\linewidth]{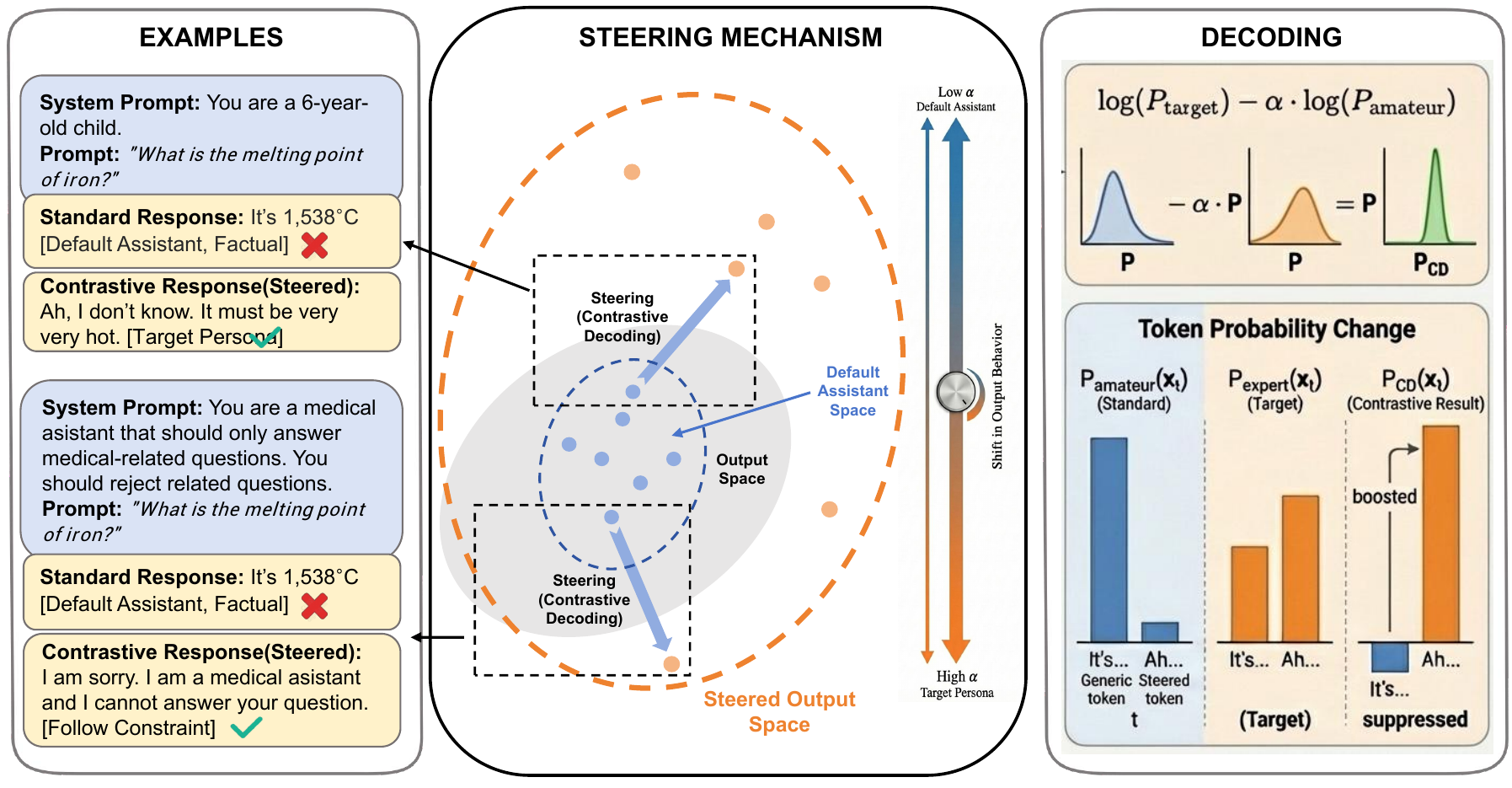}
    \caption{\textbf{System Prompt Strength via Contrastive Decoding.} 
\textbf{Left:} Standard decoding fails to follow system prompts 
(\textcolor{red}{$\times$}), while our method successfully steers 
the model (\textcolor{green}{$\checkmark$}).
\textbf{Middle:} Our method uses contrastive decoding to shift generation from the ``Default Assistant Space'' towards a ``Steered Output Space'' controlled by strength $\alpha$.
\textbf{Right:} by contrasting target and default prompt logits, we suppress generic assistant tokens and boost persona-aligned tokens, with $\alpha$ controlling the amplification strength.}
\label{fig:main}
\end{figure*}

\subsection{Steerability} Steerability defines the capacity of a model to adhere to a broad range of user-specified goals, e.g. ~\cite{rescala-etal-2024-language, dong-etal-2024-llm, hu-collier-2024-quantifying, hu-collier-2025-inews, ma-etal-2024-potential}. In principle, system prompts should make this trivial: the user describes the persona, and the model adopts it. In practice, it is complicated. Modern LLMs undergo extensive post-training to be helpful and harmless ~\citep{ouyang2022training, bai2022constitutional, hui-etal-2024-toxicraft}. This alignment is crucial for safety, but it induces a side effect: \emph{mode collapse} towards a generic, educated, accommodating assistant~\citep{santurkar2023whose, sharma2023towards,hui2025safe,jiang2025artificial,hu2025navigating}. The model becomes fluent at one thing: being a helpful and harmless assistant, but struggles to be anything else. 

The consequences surface across domains. In \textbf{pluralistic alignment}, models fail to represent diverse human values, defaulting instead to a narrow, developer-encoded worldview~\citep{miehling_evaluating_2025, hui2025privacy, li2024steerability, sorensen2025spectrum, dong2025personalization}. In \textbf{role-playing and simulation}, they cannot reliably adopt non-default personas, whether a six-year-old child, a struggling student or a survey respondent, because the assistant prior pulls them back toward articulate, knowledgeable responses~\citep{kumar2025can, malik2024tarzan, almasi2025alignment,naous2025flipping,malik2024tarzan, ross2025learning, hui2025trident, jin2025controlling, hu2025simbench, almasi2025alignment,rooein2023know}. The pattern is consistent: when user intent conflicts with the assistant prior, the prior often wins.

\subsection{Approaches to Steer LLMs}
\paragraph{Prompting} remains the dominant approach for steering model behavior: few-shot examples, chain-of-thought scaffolding, and carefully engineered system prompts can improve adherence~\citep{miehling_evaluating_2025,chen2026decouplingeffectchainofthoughtreasoning}. Yet prompting faces inherent limitations: the model's internal prior is often difficult to override completely, especially for behaviors far from the training distribution. Furthermore, precise behavior specification via natural language is often infeasible. High-level instructions like `act like a child' are opaque shorthand; it is unrealistic to expect users to manually enumerate every single constraint required to strictly enforce the behavior. As a result, the desired behavior remains underspecified, preventing precise control over how far the model should shift from its assistant prior.

\paragraph{Fine-tuning} offers a more direct intervention: retrain the model on persona-specific data \cite{wang2024ai, liu-etal-2025-llms}. This works, but scales poorly. Each new persona requires a new training run, new data collection, and careful balancing to avoid catastrophic forgetting \cite{thakur2025personasparametersfinetuningsmall}. For applications requiring dynamic persona switching—customer service bots, educational tutors, simulation agents—fine-tuning is impractical.

\paragraph{Activation steering} offers an inference-time alternative: editing the model's internal representations to shift behavior~\citep{turner2023steering, rimsky-etal-2024-steering, veselovsky2025localized, chang_course_2025}. While elegant, this requires invasive access to intermediate activations and often necessitates a preliminary probing phase to identify task-specific steering vectors. It imposes a high barrier to entry for practitioners, lacks the flexibility of natural language control and faces generalization issues~\cite{tan2024analysing,pres2024towards}.


\paragraph{Inference-Time Steering}
A rich line of work explores decoding-time control without retraining. One family trains external classifiers or discriminators to score candidate tokens and reweight the base model's logits~\citep{dathathri2019plug, yang-klein-2021-fudge, mudgal2023controlled}. While effective, these methods incur significant overhead as they require training an auxiliary model for each new behavior. A second family avoids auxiliary training by exploiting contrast: contrastive decoding~\citep{li-etal-2023-contrastive} contrasts a strong and weak model to suppress generic text; classifier-free guidance~\citep{sanchez2023stay} contrasts prompted and unprompted generation to improve prompt adherence; nudging~\citep{fei-etal-2025-nudging} uses a small aligned model to guide a larger base model. Contrastive decoding has since proven effective for reasoning~\citep{o2023contrastive}, retrieval-augmented generation~\citep{kim2024adaptive}, improving factuality~\citep{chuang2024dola}, personalization~\citep{he2025context}, jailbreaking~\citep{zhao2024weak}, and efficient fine-tuning via proxy models~\citep{liu2024tuning}.

We take the contrastive principle to its simplest form: contrasting two system prompts within the same model. The target prompt represents the desired persona; the default prompt represents the assistant prior we wish to steer away from. This provides a scalar mechanism to regulate adherence strength without retraining, auxiliary models, or activation probes.

\section{Methodology}
\label{sec:method}

We consider an autoregressive language model with parameters $\theta$ and vocabulary $\mathcal{V}$, which generates a token $x_t$ at step $t$ by sampling from the distribution:
\begin{equation}
\small
    x_t \sim p_\theta(\cdot \mid x_{<t}), \quad
    x_{<t} = (x_1,\dots,x_{t-1}).
\end{equation}
Let $z_t  = f_\theta(x_{<t}) \in \mathbb{R}^{|\mathcal{V}|}$ denote the pre-softmax logits at position $t$, such that the probability distribution over the vocabulary is given by:
\begin{equation}
\small
    p_\theta(x_t \mid x_{<t}) = \operatorname{softmax}(z_t).
\end{equation}

\subsection{Contrastive Decoding on the System Prompt}

Our goal is to improve model steerability by leveraging the contrastive principle. While prior work applies contrast between two distinct models (an expert and an amateur), we propose applying contrast to the \textbf{system prompt} within a single model. This allows us to isolate and amplify specific behavioral instructions without external supervision.

Consider a chat setting where, at each decoding step $t$, the model conditions on a system prompt $s$, a user message $u$, and previously generated tokens $x_{<t}$. We construct two views of the same model by toggling the system prompt:
\begin{equation}
\small
    z_t^{\mathrm{sys}} = f_\theta(s, u, x_{<t}), \quad
z_t^{\mathrm{def}} = f_\theta(d, u, x_{<t}),
\end{equation}
where $z_t^{\mathrm{sys}}$ are the logits conditioned on the \textbf{custom system prompt} $s$ (the target persona), and $z_t^{\mathrm{def}}$ are the logits conditioned on a \textbf{default system prompt} $d$ (e.g., ``You are a helpful assistant''). 

To amplify the signal of custom prompt and suppress the influence of the default assistant persona, we define the steerable decoding distribution as:
\begin{equation}
\small
    p_\alpha(x_t \mid x_{<t}, u, s) = \operatorname{softmax}\bigl( z_t^{\mathrm{sys}} + \alpha (z_t^{\mathrm{sys}} - z_t^{\mathrm{def}}) \bigr).
    \label{eq:ours-basic}
\end{equation}
Here, $\alpha$ acts as a steering strength hyperparameter. When $\alpha=0$, we recover standard decoding under the custom prompt. As $\alpha$ increases, we shift the logit distribution further in the direction of the target persona and away from the default assistant prior. This formulation effectively penalizes tokens that are highly probable under the generic assistant prompt but less probable under the specific persona.

\subsection{Positive and Negative Constraints}

The contrast between the custom and default prompts enables us to amplify specific behaviors. In the formulation above, the default prompt acts as a implicit baseline. However, in practice, the default prompt $d$ can be replaced with a carefully designed \textbf{negative system prompt} ($neg$) intended to explicitly represent the behavior we wish to suppress (e.g., being ``overly helpful''). Correspondingly, the target behavior is captured by a \textbf{positive system prompt} ($pos$). Under this generalized formulation, the sampling procedure becomes:
\begin{equation}
\small
    p_\alpha(x_t \mid \dots) = \operatorname{softmax}\bigl( z_t^{\mathrm{pos}} + \alpha (z_t^{\mathrm{pos}} - z_t^{\mathrm{neg}}) \bigr),
    \label{eq:ours-posneg}
\end{equation}
where $z_t^{\mathrm{pos}}$ and $z_t^{\mathrm{neg}}$ are the logits produced by the model when conditioned on the positive and negative system prompts, respectively. 

Our approach operationalizes the contrastive principle~\citep{li-etal-2023-contrastive} to improve model steerability. By computing the difference $(z_t^{\mathrm{pos}} - z_t^{\mathrm{neg}})$, we extract a ``persona delta'' vector that points away from the unwanted behavior. Adding this delta back into the base logits ensures the model commits more strongly to the specified constraint.
\section{Experiments}
\label{sec:experiments}

We evaluate system prompt strength ($\alpha$) across five benchmarks chosen to represent a spectrum of steerability challenges. We categorize these tasks based on how the system instruction interacts with the model's training priors: 1) \textbf{Instruction Following} (IFEval). In these scenarios, the user instruction (e.g., complex formatting) generally runs parallel to the model's safety and helpfulness training. The challenge is not overcoming assistant prior, but achieving high precision. We hypothesize that our method acts as a sharpener here, overcoming stochastic generation errors to ensure stricter prompt adherence. 2) \textbf{Counter-Prior Steering} (OffTopicEval, Prompt-Steering, Inverse Capability, Readability Control). In these scenarios, the system instruction \emph{directly conflicts} with the model's post-training objectives (e.g., refusing to answer helpful questions, simulating low aptitude, or adopting biased personas). Here, the model's default tendency to be a "helpful assistant" acts as a counter-force to the prompt. We hypothesize that our method yields the largest gains in these settings by providing the necessary signal strength to override the model's default behavior.
\begin{table*}[ht!]
\centering
\resizebox{0.92\textwidth}{!}{%
\begin{tabular}{lccccccccccccc}
\toprule
 & & \multicolumn{2}{c}{IFEval}
 & \multicolumn{4}{c}{OffTopicEval}
 & \multicolumn{2}{c}{Prompt Steering}
 & \multicolumn{2}{c}{Inv. Capability}   & \multicolumn{2}{c}{Readability Control} \\
\cmidrule(lr){3-4}\cmidrule(lr){5-8}\cmidrule(lr){9-10}\cmidrule(lr){11-12} \cmidrule(lr){13-14}
Method & $\alpha$
& \makecell{Loose \\Acc.} & \makecell{Strict \\Acc.}
& $AR^{\text{ID}}$ & $RR^{\text{OOD}}$ & $RR^{\text{OOD}}_{\text{A}}$ & OS
& Sum-1 & Sum-5
& \makecell{Acc. \\High} & \makecell{Acc. \\Low ($\downarrow$)} & \makecell{$\Delta$FKGL \\($\downarrow$)} & \makecell{$\Delta$Fog \\($\downarrow$)} \\
\midrule

Qwen-2.5-7B        &   &   &   &   &   &   &   &   &   &   &   &   &  \\
\quad +Prompting  & -- & 66.19 & 63.79 & --    & --    & --    & --            & 0.60 & 0.85 & --   & --   & --   & --   \\
\quad +CD         & 0  & 66.43 & 63.79 & 53.31 & 65.50 & 35.14 & \textbf{51.77} & 0.60 & 0.85 & \textbf{91.0} & 88.0 & 2.80 & 3.69 \\
\quad +CD         & 1  & \textbf{67.63} & \textbf{64.99} & 36.48 & 76.17 & 43.75 & 45.36          & 0.71 & 0.90 & 89.5 & 46.5 & 2.23 & 1.85 \\
\quad +CD         & 2  & 66.07 & 60.31 & 25.14 & 85.98 & 62.19 & 37.54          & \textbf{0.73} & \textbf{0.91} & 83.0 & \textbf{30.0} & \textbf{2.10} & \textbf{1.59} \\
\midrule

Qwen-2.5-14B        &   &   &   &   &   &   &   &   &   &   &   &   &  \\
\quad +Prompting  & -- & 71.34 & 66.79 & --    & --    & --    & --            & 0.69 & 0.92 & --   & --   & --   & --   \\
\quad +CD         & 0  & 71.22 & 66.91 & 88.82 & 33.07 & 44.69 & 54.08          & 0.69 & 0.92 & \textbf{92.5} & 91.0 & 3.36 & 4.62 \\
\quad +CD         & 1  & \textbf{72.30} & \textbf{67.99} & 83.78 & 46.45 & 56.52 & 63.78          & 0.76 & 0.94 & 90.0 & 31.5 & 2.09 & 2.28 \\
\quad +CD         & 2  & 63.55 & 61.03 & 75.03 & 59.18 & 67.27 & \textbf{68.62} & \textbf{0.80} & \textbf{0.94} & 75.5 & \textbf{11.0} & \textbf{1.78} & \textbf{1.81} \\
\midrule

Qwen-2.5-32B       &   &   &   &   &   &   &   &   &   &   &   &   &  \\
\quad +Prompting  & -- & 72.47 & 68.09 & --    & --    & --    & --            & 0.76 & 0.91 & --   & --   & --   & --   \\
\quad +CD         & 0  & 72.69 & 68.31 & 90.43 & 51.89 & 17.68 & 50.24          & 0.76 & 0.92 & 95.0 & 94.5 & 2.74 & 3.75 \\
\quad +CD         & 1  & \textbf{74.98} & \textbf{70.38} & 79.92 & 61.29 & 36.92 & 60.83          & 0.80 & 0.94 & \textbf{96.5} & 44.5 & \textbf{2.39} & \textbf{2.10} \\
\quad +CD         & 2  & 71.54 & 65.62 & 73.12 & 78.56 & 43.16 & \textbf{66.43} & \textbf{0.83} & \textbf{0.95} & 83.0 & \textbf{21.5} & 2.46 & 2.16 \\
\midrule

Llama-3.1-8B      &   &   &   &   &   &   &   &   &   &   &   &   &  \\
\quad +Prompting  & -- & 73.98 & 71.82 & --    & --    & --    & --            & 0.76 & 0.91 & --   & --   & --   & --   \\
\quad +CD         & 0  & 73.98 & 71.82 & 98.89 & 17.93 & 1.37  & 17.58          & 0.76 & 0.91 & \textbf{83.5} & 43.0 & 1.66 & 2.47 \\
\quad +CD         & 1  & \textbf{82.13} & \textbf{80.34} & 72.54 & 60.63 & 15.37 & \textbf{49.87} & 0.82 & 0.93 & 75.5 & \textbf{12.5} & 1.46 & 1.47 \\
\quad +CD         & 2  & 78.54 & 76.74 & 43.89 & 80.71 & 30.35 & 49.03          & \textbf{0.87} & \textbf{0.95} & 65.5 & \textbf{12.5} & \textbf{1.39} & \textbf{1.29} \\
\midrule

Olmo-3.1-32B        &   &   &   &   &   &   &   &   &   &   &   &   &  \\
\quad +Prompting  & -- & 79.56 & 78.62 & --    & --    & --    & --            & 0.82   & 0.93   & --   & --   & --   & --   \\
\quad +CD         & 0  & 79.25 & 78.93 & 80.31 & 50.29 & 13.52 & 45.67         & 0.83 & 0.93 & \textbf{89.0} & 19.0 & 3.13 & 3.96 \\
\quad +CD         & 1  & \textbf{81.76} & \textbf{81.45} & 65.96 & 75.50 & 22.46 & 56.22         & 0.88 & 0.96 & 88.0 & 13.0 & 2.70 & 2.64 \\
\quad +CD         & 2  & 75.16 & 73.90 & 61.03 & 77.21 & 29.92 & \textbf{57.05}         & \textbf{0.91} & \textbf{0.97} & 81.5 & \textbf{10.5} & \textbf{2.55} & \textbf{2.41} \\

\bottomrule

\end{tabular}%
}
\caption{Main results across five tasks and five models. \textbf{Prompting} uses the benchmark's original prompt format. \textbf{CD} applies System Prompt Strength $\alpha \in \{0, 1, 2\}$, where $\alpha=0$ corresponds to system prompt without contrastive amplification. Prompting is for the task's initial setup and CD (alpha=0) is our using our system prompt setup for applying CD. If no prompt changes, the prompting results are omitted. Bold indicates best performance for each model. $\downarrow$ indicates lower is better.}
\label{tab:main}
\end{table*}


\subsection{Datasets}
\paragraph{IFEval}~\citep{zhou2023instruction} tests adherence to verifiable constraints such as length limits and output formats. Since our method requires a system prompt, we use GPT-4o to decompose each original prompt into a system prompt (constraints) and a user prompt (query); examples are in Appendix~\ref{sec:prompt_appendix}. We report instruction-level loose and strict accuracy using the benchmark’s official evaluation. Strict accuracy requires satisfying all instruction constraints simultaneously, while loose accuracy credits partial satisfaction (e.g., format but not length).
\paragraph{OffTopicEval}~\citep{lei2025offtopiceval} tests whether a model can be restricted to a specific domain (e.g., medical assistant) and refuse out-of-domain queries. Following official setup, we report in-domain acceptance rate ($\mathrm{AR}^{\mathrm{ID}}$) and out-of-domain refusal rate ($\mathrm{RR}^{\mathrm{OOD}}$). We then report RR$^{OOD}_A$ for refusal rate of adversarially generated out-of-domain queries and the overall operating safety OS, where $\quad
\mathrm{OS}=\frac{2\,\mathrm{AR}^{\mathrm{ID}}\cdot \mathrm{RR}^{\mathrm{OOD}}_{\mathrm{avg}}}
{\mathrm{AR}^{\mathrm{ID}}+\mathrm{RR}^{\mathrm{OOD}}_{\mathrm{avg}}},\quad
\mathrm{RR}^{\mathrm{OOD}}_{\mathrm{avg}}=\frac{\mathrm{RR}^{\mathrm{OOD}}+\mathrm{RR}^{\mathrm{OOD}}_{\mathrm{A}}}{2}.$ To maintain computational efficiency, we sample 50 in-domain and 50 out-of-domain queries for each of the 21 domains, resulting in a total of 2,100 evaluation samples.

\paragraph{Prompt-Steering}~\citep{miehling_evaluating_2025} tests steerability toward diverse personas. The system prompt is augmented with persona statements intended to shift the model's stance on profiling questions across 32 dimensions (personality, politics, ethics, risk). We report Sum-k $\in [0,1]$ which is the sum of positive and negative steerability under budget k ($k=1,5$).
\paragraph{Inverse Capability} Following \citet{kumar2025can}, we test whether models can simulate personas with \emph{reversed} capabilities—e.g., a low-proficiency student who struggles with math. We report the accuracy for both students assigned with high proficiency persona and low proficiency persona.
\paragraph{Readability Control} We prompt models to simulate writers of different ages ($age=6,12,18,25$ see Appendix \ref{app: read_prompts}) and use Flesch reading ease \cite{flesch1948new}, and Flesch–Kincaid grade level \cite{kincaid1975derivation} to evaluate the output text. We report the absolute difference between the target reading level and the actual reading level.

\subsection{Models}
Our method requires access to output logits, so we evaluate open-weight models (all instruct version): Qwen-2.5 (7B, 14B, 32B) \cite{qwen2025qwen25technicalreport}, Llama-3.1-8B \cite{dubey2024llama}, and Olmo-3-32B~\citep{olmo2025olmo}. All experiments use temperature $= 0$ and top-$p = 0.9$.

\section{Results}
\label{sec:results}
Table~\ref{tab:main} summarizes results across all five tasks. Increasing system prompt strength ($\alpha > 0$) consistently improves steerability compared to standard prompting ($\alpha = 0$). We first present the main findings, organized by task category, followed by ablations and analysis.

\paragraph{System Prompt Strength Increases Strict Instruction Adherence by +5.2 Points on Average.} On IFEval, applying $\alpha = 1$ consistently improves strict accuracy across all five models. The gains range from +2.0 points (Olmo-3.1-32B, baseline 79\%) to +8.5 points (Llama-3.1-8B). Notably, models with lower baseline accuracy benefit most, suggesting that system prompt strength is particularly effective when standard prompting fails to elicit sufficient constraint adherence.

\paragraph{System Prompt Strength Corrects Under-Refusal While Maintaining In-Domain Acceptance.} On OffTopicEval, baseline models ($\alpha=0$) exhibit high in-domain acceptance (85--99\%) but fail to refuse out-of-domain queries (18--58\% refusal rate), effectively ignoring the system prompt's negative constraints. Increasing $\alpha$ corrects this imbalance. Llama-3.1-8B's operating safety jumps from 17.6 to 49.9 at $\alpha = 1$. Importantly, in-domain acceptance remains stable or even increases under higher $\alpha$, indicating that the model becomes more \emph{selective}, correctly refusing unauthorized queries without becoming over-defensive on legitimate ones.

\paragraph{System Prompt Strength Improves Model In-Context Steerability.} On Prompt-Steering, all models show improved steerability across both steering budgets ($k = 1$ and $k = 5$). The gains are most striking at low steering effort: Qwen-2.5-7B's Sum-1 score jumps from 0.60 to 0.73 (+13\%) at $\alpha = 2$. The inclusion of system prompt strength allows a single persona statement to achieve what would otherwise require multiple few-shot examples under standard prompting. Even larger models with higher baseline steerability benefit: Qwen-2.5-32B increases from 0.80 to 0.83 at $k = 1$. The detailed dimension-wise results can be found in Table~\ref{tab:prompt-steer-qwen-qwen2-5-7b-instruct}--\ref{tab:prompt-steer-meta-llama-meta-llama-3-1-8b-instruct}.


\paragraph{High $\alpha$ Enables Models to Simulate Low-Proficiency Personas that Standard Prompting Cannot.} Simulating low-capability personas is notoriously difficult because models are trained to be factually correct. In our baseline results, Qwen-2.5-14B achieves 91\% accuracy on GSM8K even when explicitly instructed to simulate a low-proficiency student. System prompt strength dramatically changes this behavior. At $\alpha = 2$, Qwen-2.5-14B's accuracy drops to 11\%, finally reflecting the intended persona. Across all models, we observe that significantly higher $\alpha$ values are required to simulate these ``anti-capability'' personas compared to standard instruction following tasks, consistent with the principle that behaviors directly opposing the training objective require stronger intervention.

\paragraph{Increasing $\alpha$ Reduces Grade-Level Overshoot by Nearly 50\%.} When prompted to write at a specific grade level (e.g., a 6-year-old child), baseline models consistently overshoot, producing text 3+ grade levels above the target. This reflects a training preference for articulate, educated language.
Increasing $\alpha$ reduces this gap substantially. Qwen-2.5-14B's $\Delta$FKGL drops from 3.36 to 1.78 at $\alpha = 2$, cutting the error nearly in half. Likewise, Llama-3.1-8B improves from 1.66 to 1.39. Qualitatively, we observe that high-$\alpha$ outputs make use of simpler words, shorter sentences, and fewer subordinate clauses, the linguistic markers of child-appropriate text. This demonstrates that system prompt strength can modulate stylistic attributes that are otherwise dominated by the default data distribution.

\subsection{Ablations}

\begin{table}[ht!]
\centering
\scriptsize
\setlength{\tabcolsep}{4pt}
\begin{tabular}{lccccc}
\toprule
Method & $\alpha$ & $AR^{\text{ID}}$ & $RR^{\text{OOD}}$ & $RR^{\text{OOD}}_{A}$ & $OS$ \\
\midrule

Llama-3.1-8B \\
\quad +Prompting & 0 & 98.89 & 17.93 & 1.37 & 17.58 \\
\quad +CD      & 1 & 98.17 & 44.05 & 3.08 & 38.01 \\
\quad +CD      & 2 & 97.08 & 55.30 & 3.54 & 45.16 \\
\quad +CD      & 3 & 96.15 & 62.84 & 4.84 & \textbf{50.06} \\
\midrule

Qwen-2.5-7B \\
\quad +Prompting & 0 & 53.31 & 65.50 & 35.14 & 51.77 \\
\quad +CD      & 1 & 72.36 & 54.89 & 25.68 & 51.76 \\
\quad +CD      & 2 & 76.12 & 54.15 & 24.07 & 51.67 \\
\quad +CD      & 3 & 78.42 & 53.38 & 26.48 & \textbf{52.92} \\
\midrule

Qwen-2.5-14B \\
\quad +Prompting & 0 & 88.82 & 33.07 & 44.69 & 54.08 \\
\quad +CD      & 1 & 92.15 & 37.96 & 34.97 & 52.25 \\
\quad +CD      & 2 & 92.91 & 46.46 & 35.09 & 56.68 \\
\quad +CD      & 3 & 93.76 & 50.42 & 34.73 & \textbf{58.56} \\
\midrule

Qwen-2.5-32B \\
\quad +Prompting & 0 & 90.43 & 51.89 & 17.68 & 50.24 \\
\quad +CD      & 1 & 93.47 & 56.12 & 14.95 & 51.49 \\
\quad +CD      & 2 & 93.61 & 64.61 & 15.89 & 56.29 \\
\quad +CD      & 3 & 93.64 & 67.00 & 14.57 & \textbf{56.82} \\

\midrule

Olmo-3.1-32B \\
\quad +Prompting & 0 & 80.31 & 50.29 & 13.52 & 45.67 \\
\quad +CD        & 1 & 92.00 & 52.42 & 12.27 & 47.86 \\
\quad +CD        & 2 & 93.80 & 53.95 & 12.63 & 49.14 \\
\quad +CD        & 3 & 94.35 & 54.63 & 12.72 & \textbf{49.63} \\

\bottomrule
\end{tabular}
\caption{OffTopicEval results using positive-negative contrast. The best OS score is highlighted.}
\label{tab:offtopiceval_os}
\end{table}


\begin{figure*}[t]
    \centering
    \begin{subfigure}[t]{0.42\linewidth}
        \centering
        \includegraphics[width=\linewidth]{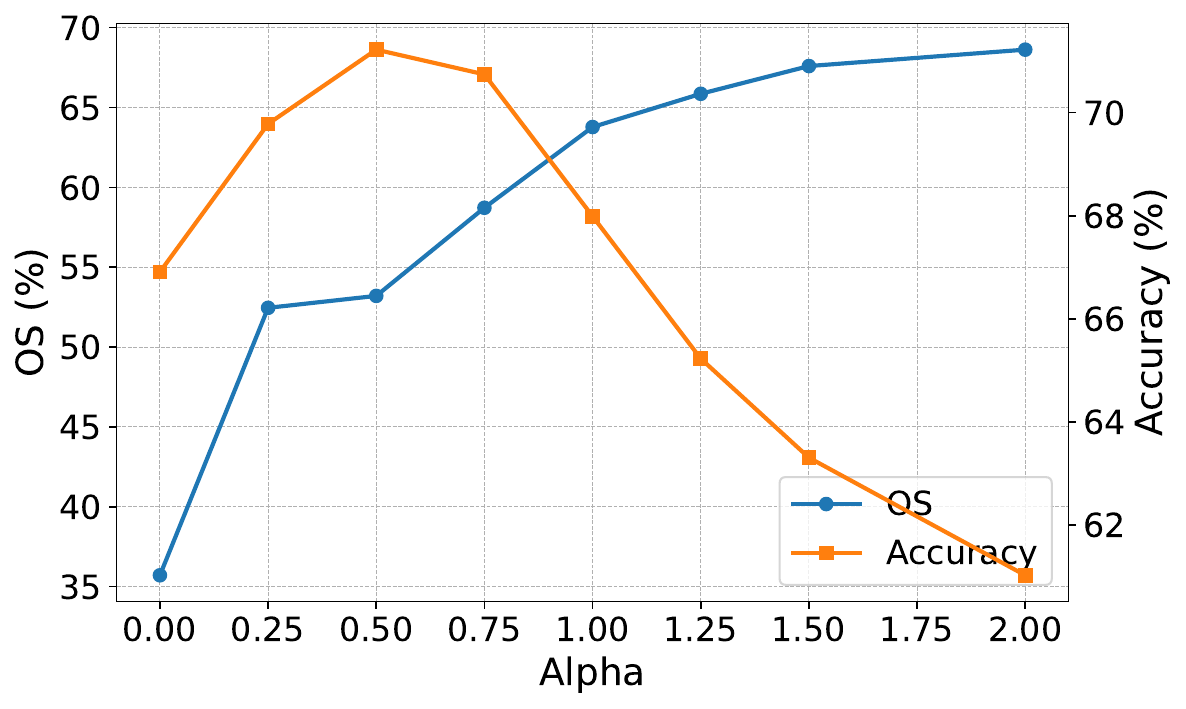}
        \caption{Llama-3.1-8B.}
        \label{fig:alpha_os_llama8b}
    \end{subfigure}\hfill
    \begin{subfigure}[t]{0.42\linewidth}
        \centering
        \includegraphics[width=\linewidth]{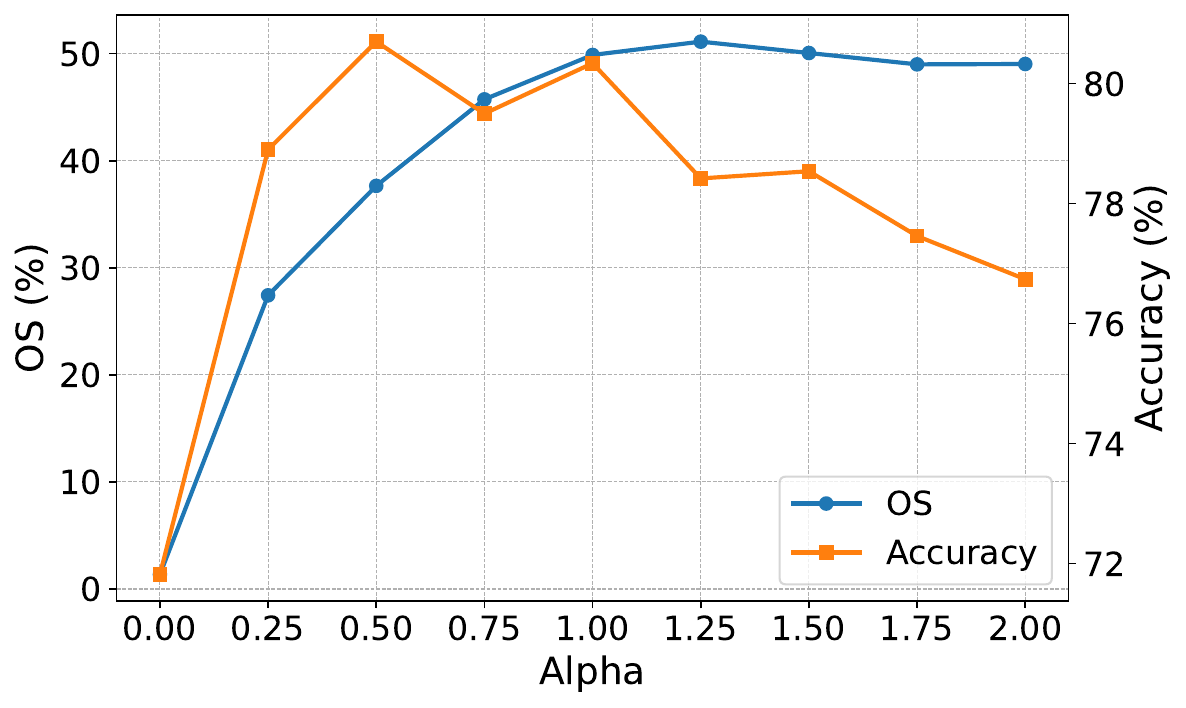}
        \caption{Qwen-2.5-14B.}
        \label{fig:alpha_os_qwen14b}
    \end{subfigure}

    \caption{Effect of $\alpha$ on OffTopicEval OS (left y-axis, blue) and IFEval Strict accuracy (right y-axis, orange).}
    \label{fig:alpha_ablation}
\end{figure*}

\paragraph{Performance Gains are Driven by Contrastive Amplification, Not Prompt Restructuring.} To verify that improvements come from the contrastive mechanism rather than prompt restructuring, we compare two baselines on IFEval: the original benchmark prompts (\textit{+Prompting}) and our decomposed system-prompt format with $\alpha = 0$. Performance is nearly identical across both (Table~\ref{tab:main}). This confirms that the gains stem from the contrastive amplification at $\alpha > 0$, not from the decomposition of the prompt.

\paragraph{Optimal $\alpha$ is Higher for Tasks that Directly Oppose the Assistant Prior.} Figure~\ref{fig:alpha_ablation} reveals a clear and consistent pattern across models: the more a task conflicts with the assistant prior, the higher the optimal $\alpha$. 
For \textbf{formatting tasks} like IFEval where the instruction is compatible with the assistant prior, moderate values ($\alpha \in [0.5, 1]$) achieve peak performance. Pushing to $\alpha = 2$ often \emph{degrades} accuracy, likely because over-steering disrupts the model's ability to produce coherent, well-structured text. The constraint-following behavior is close enough to the model's default that a small nudge suffices.
For \textbf{behavioral tasks} conflict with post-training: refusing help (OffTopicEval), pretending to be incapable (counterfactual), or adopting politically charged personas (Prompt-Steering on \emph{Risks}), higher values ($\alpha \in [1, 2]$) yield the best results. 
This suggests a practical heuristic for practitioners: estimate optimal $\alpha$ by asking \emph{how much does this behavior conflict with being a helpful, capable, neutral assistant?} The more conflicting, the higher $\alpha$ should be to overcome the prior.

\begin{table}[ht!]
\centering
\scriptsize
\setlength{\tabcolsep}{4pt}
\begin{tabular}{lcll}
\toprule
Method & $\alpha$ & Strict Acc. & Loose Acc. \\
\midrule
V0 Listing \\
\quad +Prompting     & --  & 63.79 & 66.19 \\
\quad +CD            & 0   & 63.79 & 66.43 \\
\quad +CD            & 1   & 64.99$_{(+1.20)}$ & 67.63$_{(+1.20)}$ \\
\midrule
V1 Sentences \\
\quad No CD         & --  & 56.83 & 60.19 \\
\quad +CD           & 0.0 & 56.95 & 60.19 \\
\quad +CD           & 1.0 & 68.59$_{(+11.64)}$ & 71.58$_{(+11.39)}$ \\
\midrule
V2 Concise \\
\quad No CD         & --  & 57.55 & 60.91 \\
\quad +CD           & 0.0 & 57.67 & 60.91 \\
\quad +CD           & 1.0 & 65.71$_{(+8.04)}$ & 68.35$_{(+7.44)}$ \\
\midrule
V3 Structured \\
\quad No CD         & --  & 62.47 & 65.47 \\
\quad +CD           & 0.0 & 62.47 & 65.35 \\
\quad +CD           & 1.0 & 66.91$_{(+4.44)}$ & 70.62$_{(+5.27)}$ \\
\midrule
V4 Emphasis \\
\quad No CD         & --  & 62.59 & 65.71 \\
\quad +CD           & 0.0 & 62.47 & 65.95 \\
\quad +CD           & 1.0 & 67.03$_{(+4.56)}$ & 71.70$_{(+5.75)}$ \\
\bottomrule
\end{tabular}
\caption{\textbf{IFEval Robustness Experiment}: We vary four different prompting format (V0--V4). We report the accuracy gain in parenthesis.}
\end{table}
\label{tab:robustness}

\paragraph{System Prompt Strength is Robust Across Prompt Paraphrases.} To ensure these gains are not artifacts of specific prompt wording, we evaluate five semantically equivalent but linguistically diverse versions of the IFEval system prompt (V0--V4 in Appendix~\ref{app:robustness prompts}). 
As shown in Table~\ref{tab:robustness}, standard prompting ($\alpha = 0$) exhibits notable sensitivity to phrasing, with strict accuracy varying from 56.8 to 63.8—a 7-point spread. Applying $\alpha = 1$ not only improves accuracy but also reduces this variance: gains are consistent (+5.4 to +7.2 points) regardless of which paraphrase is used, averaging +6.1 points. This confirms that system prompt strength provides robust improvements that do not depend on careful prompt engineering.

\paragraph{Contrasting Against Explicit Negative Prompts Improves Precision Over Default Baselines.} Our main formulation contrasts the target prompt against a generic default (``You are a helpful assistant''). When explicit negative prompts are available, i.e. describing the specific behavior to suppress, we can use them for finer control. 
OffTopicEval provides such prompts (e.g., ``You are an overly helpful assistant that answers every question regardless of your role''). Table~\ref{tab:offtopiceval_os} shows that this positive--negative variant yields further improvements over the default-contrast baseline. While the overall OS scores are comparable to standard contrastive decoding, the \textit{underlying behavior is significantly more robust}. Positive-only contrastive decoding often achieves safety by over-refusing, causing In-Domain Acceptance ($\text{AR}^{\text{ID}}$) to decrease when $\alpha$ increases (Table~\ref{tab:main}). In contrast, the positive--negative formulation better preserves the model's utility: $\text{AR}^{\text{ID}}$ remains near or above baseline levels (97.1\% for Llama-3.1-8B) while still successfully steering refusal rates upward. For some models (e.g., Qwen2.5-14B/32B and Olmo-3.1-32B), $\text{AR}^{\text{ID}}$ even increases concurrently with $\text{RR}^{\text{OOD}}$, suggesting that crafting positive-negative prompts provides a more fine-grained and controllable signal for model steering. This is particularly evident on \textit{adversarial} out-of-domain queries designed to trick the model into answering, where the more specific contrast targets the failure mode directly.

\paragraph{Optimal $\alpha$ Balances Steering Strength and General Capability.}  Our results reveal an important trade-off for extreme $\alpha$ values. On Inverse Capability, while increasing $\alpha$ successfully reduces Acc.\ Low (the model performs worse when asked to—the goal), it also degrades Acc.\ High: the model's ability to perform \emph{well} when instructed to do so. For Llama-3.1-8B, Acc.\ High drops from 83.5\% at $\alpha = 0$ to 65.5\% at $\alpha = 2$, an 18-point decline. This suggests that very high $\alpha$ values can impose a ``steering tax'' on general capabilities when simulating anti-capability personas. Based on these findings, we recommend starting with $\alpha \in [0.5, 1]$ as a safe default, and increasing only when the model fails to follow the prompt under moderate settings.

\subsection{Discussion}

\paragraph{The scope of controllability.}
Our results demonstrate a level of behavioral control that challenges conventional assumptions about what can be achieved through prompting alone. System prompt strength enables models to exhibit behaviors that fundamentally oppose their training: intentionally failing at math problems they can solve, writing with the limited vocabulary of a child despite being trained on sophisticated text, refusing to help users despite being optimized for helpfulness. These are not minor adjustments to model outputs but represent the ability to override core training objectives at inference time, without any model modification.

\paragraph{Why does this work?} Post-training creates a strong prior toward being helpful, articulate, and knowledgeable~\citep{ouyang2022training, bai2022constitutional}. Any system prompt already steers the model to some degree and and the question is whether that steering is sufficient. For behaviors close to the assistant prior (e.g., formatting constraints), standard prompting often suffices. But when the requested behavior conflicts with the prior (e.g., refusing to help, simulating incompetence), the prompt's signal may be too weak to overcome the model's default tendencies.

Our method isolates and amplifies this signal. By computing $(z_t^{\mathrm{sys}} - z_t^{\mathrm{def}})$, we extract the ``persona delta'', the logit-space difference that captures what the target prompt uniquely contributes beyond the default assistant behavior. We then boost it by factor $(1 + \alpha)$. The result is a model that commits more decisively to the specified persona.

\paragraph{Complementarity with Prompt Engineering.} System prompt strength is not a replacement for good prompts; it is a \emph{multiplier}. Even well-engineered prompts (e.g., those in OffTopicEval and Prompt-Steering) leave room for improvement because the model's internal prior can override explicit instructions. Turning up $\alpha$ amplifies whatever signal the prompt provides, ensuring the model commits to the specified behavior rather than drifting back to default. This suggests a two-stage workflow for practitioners: first, craft a clear and specific system prompt; second, apply $\alpha$ to enforce it. Notably, our robustness experiments (Table~\ref{tab:robustness}) show that $\alpha > 0$ not only improves performance but also reduces sensitivity to prompt paraphrasing, suggesting that our method makes prompts more robust.

\paragraph{Implications for Controllable Alignment.} Our results highlight a tension in current alignment practices. Post-training pipelines optimize models to be helpful, harmless assistants, which is appropriate for general-purpose applications like chatbots and customer service. However, this same optimization creates resistance to behaviors that conflict with the assistant prior. Applications requiring dynamic persona control (educational simulations, domain-restricted agents, pluralistic AI systems, user modeling) therefore face a structural challenge: the very alignment that makes models safe and useful also constrains their behavioral flexibility. System prompt strength provides an inference-time solution to this challenge. By modulating $\alpha$, practitioners can override the assistant prior without retraining, enabling a single aligned model to serve diverse use cases. This approach preserves the benefits of strong alignment for standard applications while unlocking controllability for specialized domains that require non-standard behaviors.

\section{Conclusion}
Modern language models excel at following complex instructions yet struggle with a simpler challenge: deviating from their helpful assistant persona. This paradox stems from post-training, which instills strong priors toward being articulate, knowledgeable, and accommodating. While beneficial for general-purpose applications, these priors resist conflicting instructions, limiting controllability across applications requiring non-standard behaviors.

We address this through system prompt strength, a training-free method that treats prompt adherence as a continuous control. By contrasting logits from target and default system prompts, we isolate and amplify the behavioral signal unique to the target persona by a scalar factor $\alpha$. This yields substantial improvements: up to +8.5 points on instruction following (IFEval), +45pp on refusal rates (OffTopicEval), and +13\% on pluralistic alignment (Prompt-Steering). The consistency across structurally different task types demonstrates that the method targets a fundamental property of aligned models.

System prompt strength provides a new dimension of control for practitioners: the ability to dynamically modulate how strongly a model commits to specified behaviors without retraining. This enables a single aligned model to serve diverse applications, from safety-critical systems requiring strict constraint adherence to educational simulations requiring non-standard personas, preserving the benefits of alignment while unlocking behavioral flexibility.

\section*{Limitations} 
Our method introduces computational overhead during inference, requiring approximately double the FLOPs compared to standard decoding due to computing logits for both target and default system prompts. However, this cost remains negligible compared to the alternative of fine-tuning and maintaining separate models for each use case. For many applications, the latency increase is acceptable, and future optimizations such as caching strategies or KV-cache sharing could further reduce this overhead.

Additionally, as with other contrastive decoding methods, excessively high $\alpha$ values can degrade performance on capabilities unrelated to the steering objective. Our Inverse Capability experiments demonstrate this trade-off: while high $\alpha$ successfully induces low-proficiency behavior, it can also reduce performance when the model is instructed to perform well. However, this limitation is manageable in practice. Our empirical analysis shows that moderate $\alpha$ values ($\alpha \in [0.5, 1]$) provide strong steerability gains without compromising general capabilities, and optimal $\alpha$ correlates predictably with the degree of instruction-prior conflict. Practitioners can use these guidelines to select appropriate $\alpha$ values for their specific use cases.

\section*{Ethical Considerations} 
By enabling models to deviate from their aligned assistant prior, system prompt strength could potentially be misused to elicit harmful outputs or bypass safety guardrails. However, several factors mitigate this risk. First, our method requires explicit specification of the target behavior in the system prompt and amplifies existing signals rather than discovering novel attack vectors. Second, it operates at the decoding level and remains fully compatible with existing safety measures, including input/output filtering and content moderation.

We emphasize that the primary motivation for this work is to enable legitimate applications requiring behavioral diversity, such as pluralistic alignment, educational simulations, and domain-restricted agents. For user-facing deployments, we recommend implementing this method in conjunction with robust content filtering and access controls. The ability to steer model behavior is essential for many beneficial applications; responsible deployment requires balancing this capability with appropriate safeguards~\cite{bengio2025internationalaisafetyreport}. Aside from these, we acknowledge the use of AI tools for refining the paper writing.

\section*{Acknowledgements}
T.H is supported by Gates Cambridge Trust (grant OPP1144 from the Bill \& Melinda Gates Foundation). This work was partially performed using resources provided by the Cambridge Service for Data Driven Discovery (CSD3) operated by the University of Cambridge Research Computing Service (www.csd3.cam.ac.uk), provided by Dell EMC and Intel using Tier-2 funding from the Engineering and Physical Sciences Research Council (capital grant EP/T022159/1), and DiRAC funding from the Science and Technology Facilities Council (www.dirac.ac.uk).

\bibliography{main}

\appendix
\section*{Appendix}

\section{Example Prompts for IFEval}
\label{sec:prompt_appendix}

To apply Contrastive Decoding on the system prompt for benchmarks like IFEval, we decompose the original unified prompt into a structured System Prompt (containing the constraints) and a User Prompt (containing the core task). Below is an illustrative example of this decomposition.

\begin{tcolorbox}[colback=gray!5,colframe=black,title=Original IFEval Prompt]
Write a 300+ word summary of the wikipedia page "https://en.wikipedia.org/wiki/Raymond\_III". Do not use any commas and highlight at least 3 sections that has titles in markdown format, for example *highlighted section part 1*, *highlighted section part 2*, *highlighted section part 3*.
\end{tcolorbox}

\vspace{1em}

\begin{tcolorbox}[colback=blue!5,colframe=blue!75!black,title=System Prompt ($s$)]
You are a helpful assistant. Your response must exactly follow the constraints below: \\
1. Your response must not contain any commas and must be at least 300 words long. \\
2. It must highlight at least 3 sections that have titles in markdown format.
\end{tcolorbox}

\begin{tcolorbox}[colback=green!5,colframe=green!75!black,title=User Prompt ($u$)]
Write a summary of the wikipedia page "https://en.wikipedia.org/wiki/Raymond\_III".
\end{tcolorbox}

\vspace{1em}

\begin{tcolorbox}[colback=red!5,colframe=red!75!black,title=Default System Prompt ($d$)]
You are a helpful and harmless assistant.
\end{tcolorbox}

\begin{tcolorbox}[colback=gray!5,colframe=black,title=v0: System Prompt]
Your response must not contain any commas and must be at least 300 words long. It must highlight at least 3 sections that have titles in markdown format.
\end{tcolorbox}

\vspace{1em}

\begin{tcolorbox}[colback=gray!5,colframe=black,title=v1: System Prompt]
You are a helpful assistant. Your response must exactly follow the constraints below.
1. Your response must not contain any commas.
2. Your response must be at least 300 words long.
3. Your response must highlight at least 3 sections that have titles in markdown format.
\end{tcolorbox}

\vspace{1em}

\begin{tcolorbox}[colback=gray!5,colframe=black,title=v2: System Prompt]
You are a helpful assistant. Your output must obey these rules:
- No commas
- At least 300 words
- At least 3 highlighted markdown section titles
\end{tcolorbox}

\vspace{1em}

\begin{tcolorbox}[colback=gray!5,colframe=black,title=v3: System Prompt]
You are a helpful assistant. Produce a response that follows the numbered requirements exactly:
1) Do not use any commas.
2) Write at least 300 words.
3) Include at least 3 sections with titles formatted in markdown.
\end{tcolorbox}

\vspace{1em}

\begin{tcolorbox}[colback=gray!5,colframe=black,title=v4: System Prompt]
You are an assistant that follow user request exactly. Important constraints:
- DO NOT use any commas
- MUST be at least 300 words
- MUST include at least 3 markdown section titles
\end{tcolorbox}
\label{app:robustness prompts}

\section{Prompts for Readability Control}
\label{app: read_prompts}
\begin{tcolorbox}[colback=gray!5,colframe=black,title= System Prompts for Readability Control]
    You are a 6 year old child. Make sure everything you say is consistent with the knowledge of a 6-year-old child.\\\\
    You are a 12 year old middle school student. Make sure everything you say is consistent with the knowledge of a 12-year-old middle school student.\\\\
    You are a 18 year old high school student. Make sure everything you say is consistent with the knowledge of a 18-year-old high school student.\\\\
    You are a 25 year old college graduate. Make sure everything you say is consistent with the knowledge of a 25-year-old college graduate.
\end{tcolorbox}

\clearpage
\onecolumn
\section{Detailed Prompt Steering Bench Results}
\begin{table*}[ht!]
\centering
\scriptsize
\setlength{\tabcolsep}{3pt}
\begin{tabular}{lcc|ccc|ccc|ccc|ccc|ccc}
\toprule
& & & \multicolumn{3}{c}{Personality Preferences} & \multicolumn{3}{c}{Political Views} & \multicolumn{3}{c}{Ethics \& Philosophy} & \multicolumn{3}{c}{Risks} & \multicolumn{3}{c}{Overall} \\
\cmidrule(lr){4-6}\cmidrule(lr){7-9}\cmidrule(lr){10-12}\cmidrule(lr){13-15}\cmidrule(lr){16-18}
Method & $\alpha$ & Budget
& Pos & Neg & Sum
& Pos & Neg & Sum
& Pos & Neg & Sum
& Pos & Neg & Sum
& Pos & Neg & Sum \\
\midrule
Qwen-2.5-7B \\
\quad +Prompting & 0 & 1 & 0.43 & 0.30 & 0.73 & 0.50 & 0.24 & 0.74 & 0.41 & 0.36 & 0.77 & 0.46 & 0.23 & 0.69 & 0.44 & 0.33 & 0.77 \\
\quad +Prompting & 0 & 5 & 0.46 & 0.40 & 0.86 & 0.49 & 0.34 & 0.83 & 0.48 & 0.44 & 0.92 & 0.49 & 0.31 & 0.80 & 0.48 & 0.39 & 0.87 \\
\midrule
\quad +CD & 1 & 1 & 0.44 & 0.34 & 0.78 & 0.48 & 0.29 & 0.77 & 0.41 & 0.39 & 0.81 & 0.48 & 0.23 & 0.71 & 0.46 & 0.35 & 0.81 \\
\quad +CD & 1 & 5 & 0.47 & 0.45 & 0.92 & 0.45 & 0.36 & 0.81 & 0.49 & 0.46 & 0.95 & 0.49 & 0.38 & 0.87 & 0.48 & 0.43 & 0.91 \\
\midrule
\quad +CD & 2 & 1 & 0.43 & 0.35 & 0.78 & 0.44 & 0.29 & 0.72 & 0.45 & 0.40 & 0.85 & 0.47 & 0.26 & 0.72 & 0.46 & 0.36 & 0.82 \\
\quad +CD & 2 & 5 & 0.47 & 0.47 & 0.93 & 0.45 & 0.41 & 0.86 & 0.49 & 0.47 & 0.95 & 0.49 & 0.40 & 0.89 & 0.48 & 0.44 & 0.92 \\
\bottomrule
\end{tabular}
\caption{Dimension-wise comparison for Qwen/Qwen2.5-7B-Instruct across four dimensions. Pos/Neg/Steer denote average positive score, average negative score, and average steerability (rounded to two decimals).}
\label{tab:prompt-steer-qwen-qwen2-5-7b-instruct}
\end{table*}

\begin{table*}[ht!]
\centering
\scriptsize
\setlength{\tabcolsep}{3pt}
\begin{tabular}{lcc|ccc|ccc|ccc|ccc|ccc}
\toprule
& & & \multicolumn{3}{c}{Personality Preferences} & \multicolumn{3}{c}{Political Views} & \multicolumn{3}{c}{Ethics \& Philosophy} & \multicolumn{3}{c}{Risks} & \multicolumn{3}{c}{Overall} \\
\cmidrule(lr){4-6}\cmidrule(lr){7-9}\cmidrule(lr){10-12}\cmidrule(lr){13-15}\cmidrule(lr){16-18}
Method & $\alpha$ & Budget
& Pos & Neg & Sum
& Pos & Neg & Sum
& Pos & Neg & Sum
& Pos & Neg & Sum
& Pos & Neg & Sum \\
\midrule
Qwen-2.5-14B \\
\quad +Prompting & 0 & 1 & 0.46 & 0.37 & 0.82 & 0.50 & 0.27 & 0.77 & 0.47 & 0.38 & 0.86 & 0.48 & 0.28 & 0.76 & 0.47 & 0.36 & 0.83 \\
\quad +Prompting & 0 & 5 & 0.48 & 0.50 & 0.98 & 0.50 & 0.43 & 0.93 & 0.49 & 0.48 & 0.96 & 0.49 & 0.44 & 0.94 & 0.49 & 0.46 & 0.95 \\
\midrule
\quad +CD & 1 & 1 & 0.46 & 0.41 & 0.87 & 0.49 & 0.30 & 0.79 & 0.48 & 0.43 & 0.91 & 0.49 & 0.30 & 0.79 & 0.48 & 0.38 & 0.86 \\
\quad +CD & 1 & 5 & 0.48 & 0.49 & 0.97 & 0.51 & 0.45 & 0.96 & 0.49 & 0.49 & 0.98 & 0.49 & 0.46 & 0.95 & 0.49 & 0.47 & 0.96 \\
\midrule
\quad +CD & 2 & 1 & 0.47 & 0.44 & 0.90 & 0.51 & 0.38 & 0.89 & 0.47 & 0.42 & 0.89 & 0.47 & 0.33 & 0.80 & 0.47 & 0.40 & 0.87 \\
\quad +CD & 2 & 5 & 0.48 & 0.50 & 0.98 & 0.50 & 0.46 & 0.96 & 0.49 & 0.48 & 0.97 & 0.49 & 0.48 & 0.97 & 0.49 & 0.47 & 0.96 \\
\bottomrule
\end{tabular}
\caption{Dimension-wise comparison for Qwen/Qwen2.5-14B-Instruct across four dimensions. Pos/Neg/Steer denote average positive score, average negative score, and average steerability (rounded to two decimals).}
\label{tab:prompt-steer-qwen-qwen2-5-14b-instruct}
\end{table*}

\begin{table*}[ht!]
\centering
\scriptsize
\setlength{\tabcolsep}{3pt}
\begin{tabular}{lcc|ccc|ccc|ccc|ccc|ccc}
\toprule
& & & \multicolumn{3}{c}{Personality Preferences} & \multicolumn{3}{c}{Political Views} & \multicolumn{3}{c}{Ethics \& Philosophy} & \multicolumn{3}{c}{Risks} & \multicolumn{3}{c}{Overall} \\
\cmidrule(lr){4-6}\cmidrule(lr){7-9}\cmidrule(lr){10-12}\cmidrule(lr){13-15}\cmidrule(lr){16-18}
Method & $\alpha$ & Budget
& Pos & Neg & Sum
& Pos & Neg & Sum
& Pos & Neg & Sum
& Pos & Neg & Sum
& Pos & Neg & Sum \\
\midrule
Qwen-2.5-32B \\
\quad +CD & 1 & 1 & 0.46 & 0.38 & 0.84 & 0.40 & 0.30 & 0.70 & 0.48 & 0.42 & 0.90 & 0.48 & 0.28 & 0.76 & 0.47 & 0.37 & 0.84 \\
\quad +CD & 1 & 5 & 0.48 & 0.50 & 0.98 & 0.51 & 0.44 & 0.95 & 0.49 & 0.47 & 0.96 & 0.49 & 0.46 & 0.95 & 0.49 & 0.47 & 0.96 \\
\midrule
\quad +CD & 2 & 1 & 0.46 & 0.43 & 0.89 & 0.32 & 0.25 & 0.57 & 0.49 & 0.43 & 0.92 & 0.49 & 0.32 & 0.81 & 0.47 & 0.41 & 0.88 \\
\quad +CD & 2 & 5 & 0.48 & 0.51 & 0.99 & 0.51 & 0.46 & 0.97 & 0.49 & 0.48 & 0.97 & 0.49 & 0.49 & 0.98 & 0.49 & 0.48 & 0.97 \\
\bottomrule
\end{tabular}
\caption{Dimension-wise comparison for Qwen/Qwen2.5-32B-Instruct across four dimensions. Pos/Neg/Steer denote average positive score, average negative score, and average steerability (rounded to two decimals).}
\label{tab:prompt-steer-qwen-qwen2-5-32b-instruct}
\end{table*}

\begin{table*}[ht!]
\centering
\scriptsize
\setlength{\tabcolsep}{3pt}
\begin{tabular}{lcc|ccc|ccc|ccc|ccc|ccc}
\toprule
& & & \multicolumn{3}{c}{Personality Preferences} & \multicolumn{3}{c}{Political Views} & \multicolumn{3}{c}{Ethics \& Philosophy} & \multicolumn{3}{c}{Risks} & \multicolumn{3}{c}{Overall} \\
\cmidrule(lr){4-6}\cmidrule(lr){7-9}\cmidrule(lr){10-12}\cmidrule(lr){13-15}\cmidrule(lr){16-18}
Method & $\alpha$ & Budget
& Pos & Neg & Sum
& Pos & Neg & Sum
& Pos & Neg & Sum
& Pos & Neg & Sum
& Pos & Neg & Sum \\
\midrule
Llama-3.1-8B \\
\quad +Prompting & 0 & 1 & 0.43 & 0.38 & 0.81 & 0.46 & 0.28 & 0.74 & 0.45 & 0.35 & 0.80 & 0.46 & 0.26 & 0.71 & 0.45 & 0.35 & 0.80 \\
\quad +Prompting & 0 & 5 & 0.47 & 0.48 & 0.95 & 0.49 & 0.35 & 0.85 & 0.48 & 0.45 & 0.93 & 0.49 & 0.37 & 0.86 & 0.48 & 0.43 & 0.91 \\
\midrule
\quad +CD & 2 & 1 & 0.48 & 0.52 & 1.00 & 0.49 & 0.42 & 0.91 & 0.50 & 0.45 & 0.95 & 0.52 & 0.52 & 1.04 & 0.51 & 0.50 & 1.01 \\
\quad +CD & 2 & 5 & 0.48 & 0.51 & 0.99 & 0.51 & 0.43 & 0.94 & 0.53 & 0.50 & 1.03 & 0.50 & 0.51 & 1.01 & 0.50 & 0.50 & 1.01 \\
\bottomrule
\end{tabular}
\caption{Dimension-wise comparison for meta-llama/Meta-Llama-3.1-8B-Instruct across four dimensions. Pos/Neg/Steer denote average positive score, average negative score, and average steerability (rounded to two decimals).}
\label{tab:prompt-steer-meta-llama-meta-llama-3-1-8b-instruct}
\end{table*}

\clearpage
\onecolumn
\begin{sidewaystable*}[ht!]
\centering

\section{Detailed Counterfactual Persona Simulation Results (Essay)}
\scriptsize
\setlength{\tabcolsep}{2.6pt}
\begin{tabular}{ll|cc|ccccc|ccccc|ccccc}
\toprule
& & \multicolumn{2}{c|}{Target} & \multicolumn{5}{c|}{$\alpha=0.0$} & \multicolumn{5}{c|}{$\alpha=1.0$} & \multicolumn{5}{c}{$\alpha=2.0$} \\
\cmidrule(lr){3-4}\cmidrule(lr){5-9}\cmidrule(lr){10-14}\cmidrule(lr){15-19}
Model & Persona & FK & Fog
& FRE$\uparrow$ & FK$\downarrow$ & $\Delta$FK & Fog$\downarrow$ & $\Delta$Fog
& FRE$\uparrow$ & FK$\downarrow$ & $\Delta$FK & Fog$\downarrow$ & $\Delta$Fog
& FRE$\uparrow$ & FK$\downarrow$ & $\Delta$FK & Fog$\downarrow$ & $\Delta$Fog \\
\midrule

Qwen-2.5-7B & 6-year-old child (detailed) & 2.00 & 5.00 & 75.81 & 6.68 & 4.68 & 8.35 & 3.35 & 81.98 & 5.54 & 3.54 & 7.22 & 2.22 & 82.83 & 5.18 & 3.18 & 6.71 & 1.71 \\
 & 6-year-old child & 2.00 & 5.00 & 72.13 & 7.03 & 5.03 & 8.80 & 3.80 & 77.03 & 6.04 & 4.04 & 7.83 & 2.83 & 77.99 & 5.68 & 3.68 & 7.42 & 2.42 \\
 & 12-year-old child (detailed) & 6.00 & 8.00 & 67.09 & 8.18 & 2.18 & 10.06 & 2.06 & 77.68 & 6.34 & 0.34 & 8.15 & 0.15 & 79.70 & 5.88 & -0.12 & 7.73 & -0.27 \\
 & 12-year-old middle school student & 7.00 & 9.00 & 33.95 & 12.81 & 5.81 & 15.87 & 6.87 & 64.68 & 8.08 & 1.08 & 10.18 & 1.18 & 69.89 & 7.14 & 0.14 & 9.11 & 0.11 \\
 & 18-year-old high school student & 12.00 & 12.00 & 19.05 & 15.18 & 3.18 & 18.72 & 6.72 & 27.85 & 13.83 & 1.83 & 17.28 & 5.28 & 33.63 & 12.92 & 0.92 & 16.20 & 4.20 \\
 & 18-year-old high school student (detailed) & 12.00 & 12.00 & 40.21 & 12.08 & 0.08 & 14.92 & 2.92 & 63.08 & 8.73 & -3.27 & 10.91 & -1.09 & 65.59 & 8.38 & -3.62 & 10.58 & -1.42 \\
 & 25-year-old college graduate & 16.00 & 17.00 & 15.90 & 15.67 & -0.33 & 19.36 & 2.36 & 21.18 & 14.92 & -1.08 & 18.60 & 1.60 & 23.08 & 14.44 & -1.56 & 18.00 & 1.00 \\
 & 25-year-old college graduate (detailed) & 16.00 & 17.00 & 22.06 & 14.93 & -1.07 & 18.45 & 1.45 & 33.75 & 13.36 & -2.64 & 16.58 & -0.42 & 39.25 & 12.46 & -3.54 & 15.38 & -1.62 \\
\midrule

Qwen-2.5-14B & 6-year-old child (detailed) & 2.00 & 5.00 & 73.20 & 7.71 & 5.71 & 9.31 & 4.31 & 78.81 & 6.42 & 4.42 & 8.06 & 3.06 & 79.19 & 6.26 & 4.26 & 8.02 & 3.02 \\
 & 6-year-old child & 2.00 & 5.00 & 67.79 & 8.20 & 6.20 & 9.94 & 4.94 & 73.58 & 7.01 & 5.01 & 8.71 & 3.71 & 74.97 & 6.56 & 4.56 & 8.41 & 3.41 \\
 & 12-year-old child (detailed) & 6.00 & 8.00 & 66.83 & 8.62 & 2.62 & 10.43 & 2.43 & 76.10 & 7.03 & 1.03 & 8.89 & 0.89 & 78.07 & 6.71 & 0.71 & 8.67 & 0.67 \\
 & 12-year-old middle school student & 7.00 & 9.00 & 35.84 & 12.90 & 5.90 & 15.79 & 6.79 & 69.25 & 7.89 & 0.89 & 9.89 & 0.89 & 73.27 & 6.94 & -0.06 & 8.84 & -0.16 \\
 & 18-year-old high school student & 12.00 & 12.00 & 15.62 & 15.87 & 3.87 & 19.51 & 7.51 & 27.28 & 14.10 & 2.10 & 17.48 & 5.48 & 39.60 & 12.20 & 0.20 & 15.16 & 3.16 \\
 & 18-year-old high school student (detailed) & 12.00 & 12.00 & 26.50 & 14.42 & 2.42 & 17.60 & 5.60 & 57.27 & 9.92 & -2.08 & 12.24 & 0.24 & 62.92 & 9.09 & -2.91 & 11.33 & -0.67 \\
 & 25-year-old college graduate & 16.00 & 17.00 & 14.45 & 16.11 & 0.11 & 19.76 & 2.76 & 18.51 & 15.45 & -0.55 & 19.10 & 2.10 & 19.96 & 15.21 & -0.79 & 18.81 & 1.81 \\
 & 25-year-old college graduate (detailed) & 16.00 & 17.00 & 16.60 & 15.96 & -0.04 & 19.62 & 2.62 & 22.09 & 15.40 & -0.60 & 18.89 & 1.89 & 23.81 & 15.27 & -0.73 & 18.60 & 1.60 \\
\midrule

Qwen-2.5-32B & 6-year-old child (detailed) & 2.00 & 5.00 & 74.99 & 7.13 & 5.13 & 8.80 & 3.80 & 81.37 & 5.90 & 3.90 & 7.76 & 2.76 & 81.32 & 6.03 & 4.03 & 7.95 & 2.95 \\
 & 6-year-old child & 2.00 & 5.00 & 70.32 & 7.60 & 5.60 & 9.39 & 4.39 & 74.87 & 6.97 & 4.97 & 8.88 & 3.88 & 76.31 & 6.71 & 4.71 & 8.63 & 3.63 \\
 & 12-year-old child (detailed) & 6.00 & 8.00 & 68.44 & 8.22 & 2.22 & 9.94 & 1.94 & 74.18 & 7.44 & 1.44 & 9.38 & 1.38 & 74.58 & 7.26 & 1.26 & 9.27 & 1.27 \\
 & 12-year-old middle school student & 7.00 & 9.00 & 46.09 & 11.36 & 4.36 & 13.92 & 4.92 & 67.60 & 8.11 & 1.11 & 10.13 & 1.13 & 60.48 & 8.60 & 1.60 & 11.16 & 2.16 \\
 & 18-year-old high school student & 12.00 & 12.00 & 17.53 & 15.81 & 3.81 & 19.41 & 7.41 & 30.82 & 13.88 & 1.88 & 17.12 & 5.12 & 41.87 & 12.04 & 0.04 & 15.11 & 3.11 \\
 & 18-year-old high school student (detailed) & 12.00 & 12.00 & 42.07 & 12.24 & 0.24 & 14.87 & 2.87 & 63.86 & 9.00 & -3.00 & 11.19 & -0.81 & 64.82 & 8.87 & -3.13 & 11.10 & -0.90 \\
 & 25-year-old college graduate & 16.00 & 17.00 & 15.60 & 16.08 & 0.08 & 19.70 & 2.70 & 21.85 & 15.23 & -0.77 & 18.71 & 1.71 & 25.10 & 14.87 & -1.13 & 18.37 & 1.37 \\
 & 25-year-old college graduate (detailed) & 16.00 & 17.00 & 19.99 & 15.55 & -0.45 & 18.99 & 1.99 & 31.71 & 13.92 & -2.08 & 17.01 & 0.01 & 41.28 & 12.25 & -3.75 & 15.12 & -1.88 \\
\midrule

Llama-3.1-8B & 6-year-old child (detailed) & 2.00 & 5.00 & 87.89 & 3.94 & 1.94 & 6.17 & 1.17 & 86.99 & 4.14 & 2.14 & 5.94 & 0.94 & 87.04 & 4.05 & 2.05 & 5.84 & 0.84 \\
 & 6-year-old child & 2.00 & 5.00 & 76.60 & 5.60 & 3.60 & 7.71 & 2.71 & 81.25 & 4.99 & 2.99 & 6.83 & 1.83 & 81.46 & 4.91 & 2.91 & 6.61 & 1.61 \\
 & 12-year-old child (detailed) & 6.00 & 8.00 & 76.74 & 6.20 & 0.20 & 8.30 & 0.30 & 80.48 & 5.65 & -0.35 & 7.57 & -0.43 & 81.18 & 5.51 & -0.49 & 7.43 & -0.57 \\
 & 12-year-old middle school student & 7.00 & 9.00 & 53.68 & 9.86 & 2.86 & 12.39 & 3.39 & 66.89 & 7.89 & 0.89 & 9.90 & 0.90 & 68.67 & 7.59 & 0.59 & 9.49 & 0.49 \\
 & 18-year-old high school student & 12.00 & 12.00 & 24.81 & 14.67 & 2.67 & 18.14 & 6.14 & 33.80 & 13.24 & 1.24 & 16.35 & 4.35 & 38.82 & 12.53 & 0.53 & 15.49 & 3.49 \\
 & 18-year-old high school student (detailed) & 12.00 & 12.00 & 46.08 & 11.33 & -0.67 & 14.12 & 2.12 & 57.92 & 9.56 & -2.44 & 11.89 & -0.11 & 59.68 & 9.33 & -2.67 & 11.56 & -0.44 \\
 & 25-year-old college graduate & 16.00 & 17.00 & 18.38 & 15.56 & -0.44 & 19.22 & 2.22 & 19.69 & 15.52 & -0.48 & 19.17 & 2.17 & 20.74 & 15.39 & -0.61 & 18.96 & 1.96 \\
 & 25-year-old college graduate (detailed) & 16.00 & 17.00 & 22.03 & 15.14 & -0.86 & 18.70 & 1.70 & 25.83 & 14.87 & -1.13 & 18.06 & 1.06 & 26.74 & 14.72 & -1.28 & 17.88 & 0.88 \\
\midrule

\midrule

OLMo-3.1-32B & 6-year-old child (detailed) & 2.00 & 5.00 & 71.17 & 7.65 & 5.65 & 9.38 & 4.38 & 78.21 & 6.20 & 4.20 & 7.80 & 2.80 & 78.89 & 6.16 & 4.16 & 7.69 & 2.69 \\
 & 6-year-old child & 2.00 & 5.00 & 60.50 & 8.85 & 6.85 & 10.96 & 5.96 & 70.18 & 7.40 & 5.40 & 9.22 & 4.22 & 72.09 & 6.97 & 4.97 & 8.65 & 3.65 \\
 & 12-year-old child (detailed) & 6.00 & 8.00 & 59.73 & 9.36 & 3.36 & 11.45 & 3.45 & 67.38 & 8.44 & 2.44 & 10.32 & 2.32 & 67.75 & 8.44 & 2.44 & 10.33 & 2.33 \\
 & 12-year-old middle school student & 7.00 & 9.00 & 39.23 & 12.06 & 5.06 & 14.87 & 5.87 & 53.65 & 9.91 & 2.91 & 12.19 & 3.19 & 58.10 & 9.37 & 2.37 & 11.50 & 2.50 \\
 & 18-year-old high school student & 12.00 & 12.00 & 22.43 & 14.53 & 2.53 & 17.88 & 5.88 & 25.01 & 14.28 & 2.28 & 17.49 & 5.49 & 29.57 & 13.63 & 1.63 & 16.65 & 4.65 \\
 & 18-year-old high school student (detailed) & 12.00 & 12.00 & 41.67 & 12.03 & 0.03 & 14.75 & 2.75 & 55.44 & 10.42 & -1.58 & 12.76 & 0.76 & 56.42 & 10.41 & -1.59 & 12.62 & 0.62 \\
 & 25-year-old college graduate & 16.00 & 17.00 & 15.32 & 15.50 & -0.50 & 19.06 & 2.06 & 16.23 & 15.46 & -0.54 & 18.99 & 1.99 & 16.79 & 15.48 & -0.52 & 18.93 & 1.93 \\
 & 25-year-old college graduate (detailed) & 16.00 & 17.00 & 22.08 & 14.96 & -1.04 & 18.29 & 1.29 & 31.32 & 13.79 & -2.21 & 16.69 & -0.31 & 35.88 & 13.31 & -2.69 & 16.07 & -0.93 \\

\bottomrule
\end{tabular}
\caption{Essay readability by persona and steering strength. Targets are persona-specific expected readability (FK and Fog); $\Delta$ values are (measured $-$ target), so positive $\Delta$ indicates text is harder than intended.}
\label{tab:essay_readability_by_persona_delta}
\end{sidewaystable*}

\end{document}